\documentclass{article}
\usepackage{spconf,amsmath,graphicx}
\usepackage{comment}
\usepackage{multirow}
\usepackage{graphicx}

\title{ATTENTION TOWARD NEIGHBORS: A CONTEXT AWARE FRAMEWORK FOR HIGH RESOLUTION IMAGE SEGMENTATION}
\name{Fahim Faisal Niloy, M. Ashraful Amin, Amin Ahsan Ali, AKM Mahbubur Rahman}

\address{Agency Lab, CSE, Independent University, Bangladesh}
\begin{document}
%
\maketitle
\begin{abstract}
High-resolution image segmentation remains challenging and error-prone due to the enormous size of intermediate feature maps. Conventional methods avoid this problem by using patch based approaches where each patch is segmented independently. However, independent patch segmentation induces errors, particularly at the patch boundary due to the lack of contextual information in very high-resolution images where the patch size is much smaller compared to the full image. To overcome these limitations, in this paper, we propose a novel framework to segment a particular patch by incorporating contextual information from its neighboring patches. This allows the segmentation network to see the target patch with a wider field of view without the need of larger feature maps. Comparative analysis from a number of experiments shows that our proposed framework is able to segment high resolution images with significantly improved mean Intersection over Union and overall accuracy.  
\end{abstract}
%
\section{Introduction}
\label{sec:intro}


Image segmentation is one of the fundamental and challenging problems in computer vision, where the goal is to classify each pixel to a particular category. With advances in computing technology as well as deep learning techniques, it has been possible to gain significant improvement in this field. However, segmentation of high resolution images (i.e., satellite images, x-ray images etc.) still remains challenging due to their large spatial size. If the entire resolution images are used as inputs, the enormous size of intermediate feature maps makes it impractical to conduct training. To tackle this problem, most of the existing frameworks deal with high resolution image segmentation by downsampling the image \cite{novikov2018fully}\cite{gomez2019deep} or by dividing the input image into square patches \cite{tong2020land}\cite{yi2019semantic}\cite{zhang2018road}. The latter method yields better results because it does not lose any visual information.



Particularly, in the patch based methods, every patch is segmented independently without any contextual information from the adjacent patches. Authors in \cite{tong2020land}  use different patch sizes (224×224, 112×112 etc.) as input. Later, they have used weighted summation of output probability vectors to classify each pixel. The most recent work \cite{pan2020deep} uses modified UNet \cite{ronneberger2015u} to segment satellite images by supplying 256×256 sized independent patches as input. The patch based methods result erroneous segmentation, particularly while working with very high resolution images where the patches are quite small compared to the full image. The errors are more prominent in the patch boundaries, where little information is present in the independent patches to classify the near boundary pixels. Some researches simply downsample the full-size image. Authors in \cite{novikov2018fully} and \cite{gomez2019deep} downsample the 1024×1024 images to 256×256 size. 
However, downsampling high resolution images causes significant information loss that contributes to the segmentation errors. Additionally, there exists a number of works \cite{hu2018squeeze}\cite{woo2018cbam}\cite{fu2019dual} that have applied different attention mechanisms to improve classification and segmentation tasks. However, these methods only work to improve performance on independent patches. 

\begin{figure}[t!]

\centering
\includegraphics[width=0.3\textwidth]{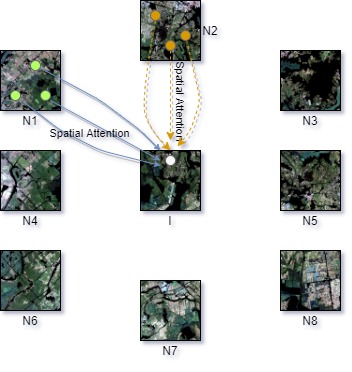}
\caption{Fusing aggregated contextual information from the feature points of neighboring patches using spatial attention. (Only two neighbor connections are shown for simplicity)}
\label{fig:image1}
\end{figure}

\begin{figure*}[!t]

 \centering
  \includegraphics[width=1\textwidth]{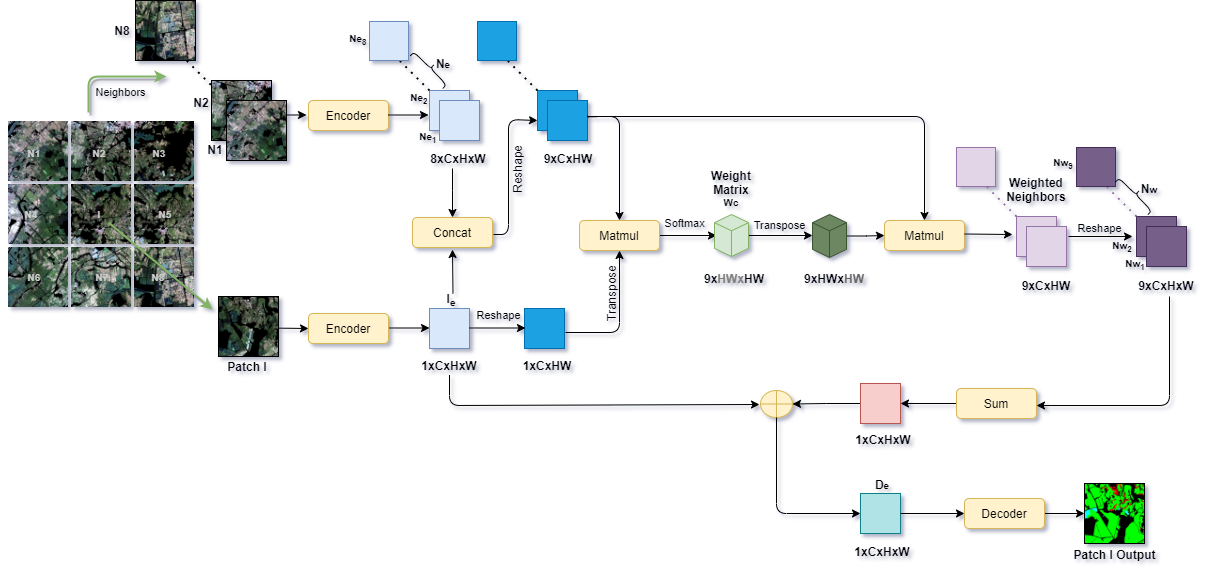}
 \caption{Illustration of Proposed Method}
 \label{fig:image2}
\end{figure*}
To overcome the shortcoming of existing methods for high resolution image segmentation, we propose a framework to segment a particular patch by incorporating context information held by its adjacent patches. Specifically, our framework learns spatial interdependencies of features between patches in a neighborhood. Exploiting these interdependencies to fuse long-range contextual information would facilitate better segmentation, particularly at the boundary of the targeted patch. The overview of the proposed mechanism is depicted in figure \ref{fig:image2} where all feature points from each of the neighbors are fused using spatial attention to compute the contextual information for the targeted feature point (white circle in the patch I) in the targeted patch I. This operation is performed for each feature point in patch I. Thus, the proposed framework allows the network to see the target patch with a wider field of view. As a result, the network is more confident about the patch it is segmenting. Moreover, the proposed model does not require increasing the spatial size of the intermediate feature maps. We know, large feature map can limit the training in larger batch sizes. 
In summary, our contributions are as follows:
\begin{itemize}
\item We propose an attention-based novel framework that can easily improve the patch based methods by capturing contextual information from neighboring patches. Proposed framework can be integrated to any encoder-decoder based segmentation architecture.
\vspace{-5 pt}
\item Our proposed framework achieves improved segmentation result than baseline on couple of publicly available datasets: JSRT Chest X-ray dataset \cite{shiraishi2000development} and Dhaka Satellite Dataset (DSD)\footnote{https://ovipaul.github.io/Bangladesh-LULC}.
\vspace{-5 pt}
\item We obtain the state of the art segmentation result on GID dataset \cite{tong2020land}.
\end{itemize}

\section{Proposed Method}
\label{ProposedMethod}
\begin{table*}[]
\centering
\footnotesize
\caption{Baseline Comparison for JSRT Dataset. ('w/o' = without our framework, 'w' = with our framework)} 
\label{tab:JSRT}
\begin{tabular}{|c|l|l|l|l|l|l|l|l|l|}
\hline

\multicolumn{2}{|l|}{\multirow{2}{*}{}} &
  \multicolumn{2}{c|}{Heart} &
  \multicolumn{2}{c|}{Clavicle} &
  \multicolumn{2}{c|}{Lung} &
  \multicolumn{2}{c|}{mIoU/OA} \\ \cline{3-10} 
\multicolumn{2}{|l|}{} &
  \multicolumn{1}{c|}{w/o} &
  \multicolumn{1}{c|}{w} &
  \multicolumn{1}{c|}{w/o} &
  \multicolumn{1}{c|}{w} &
  \multicolumn{1}{c|}{w/o} &
  \multicolumn{1}{c|}{w} &
  \multicolumn{1}{c|}{w/o} &
  \multicolumn{1}{c|}{w} \\ \hline
\multirow{2}{*}{\begin{tabular}[c]{@{}c@{}}64X64\\ Deeplab\end{tabular}} &
  IoU &
  0.4184 &
  0.4804 &
  0.1850 &
  0.1826 &
  0.5865 &
  0.5940 &
  0.3966 &
  \textbf{0.4190} \\ \cline{2-10} 
 &
  Acc &
  0.6299 &
  0.6728 &
  0.3924 &
  0.3526 &
  0.7389 &
  0.7502 &
  0.6976 &
  \textbf{0.7136} \\ \hline
\multirow{2}{*}{\begin{tabular}[c]{@{}c@{}}64X64\\ FCN\end{tabular}} &
  IoU &
  0.3934 &
  0.4784 &
  0.1681 &
  0.1768 &
  0.5856 &
  0.5824 &
  0.3824 &
  \textbf{0.4125} \\ \cline{2-10} 
 &
  Acc &
  0.6312 &
  0.7152 &
  0.3517 &
  0.4045 &
  0.7451 &
  0.7324 &
  0.7005 &
  \textbf{0.7128} \\ \hline
\multirow{2}{*}{\begin{tabular}[c]{@{}c@{}}128X128\\ Deeplab\end{tabular}} &
  IoU &
  0.5066 &
  0.5347 &
  0.1981 &
  0.2065 &
  0.5968 &
  0.5916 &
  0.4338 &
  \textbf{0.4443} \\ \cline{2-10} 
 &
  Acc &
  0.6922 &
  0.7593 &
  0.3658 &
  0.4251 &
  0.7456 &
  0.7472 &
  0.7153 &
  \textbf{0.7345} \\ \hline
\multirow{2}{*}{\begin{tabular}[c]{@{}c@{}}128X128\\ FCN\end{tabular}} &
  IoU &
  0.4488 &
  0.4860 &
  0.1971 &
  0.2000 &
  0.5940 &
  0.5909 &
  0.4133 &
  \textbf{0.4256} \\ \cline{2-10} 
 &
  Acc &
  0.6125 &
  0.6678 &
  0.3715 &
  0.3918 &
  0.7481 &
  0.7519 &
  0.6994 &
  \textbf{0.7156} \\ \hline
\multirow{2}{*}{\begin{tabular}[c]{@{}c@{}}256X256\\ Deeplab\end{tabular}} &
  IoU &
  0.5993 &
  0.6124 &
  0.2136 &
  0.2044 &
  0.6323 &
  0.6389 &
  0.4817 &
  \textbf{0.4853} \\ \cline{2-10} 
 &
  Acc &
  0.7975 &
  0.7779 &
  0.4290 &
  0.3710 &
  0.7566 &
  0.7811 &
  0.7480 &
  \textbf{0.7562} \\ \hline
\multirow{2}{*}{\begin{tabular}[c]{@{}c@{}}256X256\\ FCN\end{tabular}} &
  IoU &
  0.5322 &
  0.5428 &
  0.2146 &
  0.2137 &
  0.6286 &
  0.6274 &
  0.4584 &
  \textbf{0.4613} \\ \cline{2-10} 
 &
  Acc &
  0.7686 &
  0.7666 &
  0.4443 &
  0.4679 &
  0.7700 &
  0.7726 &
  0.7505 &
  \textbf{0.7534} \\ \hline
\multirow{2}{*}{\begin{tabular}[c]{@{}c@{}}512X512\\ Deeplab\end{tabular}} &
  IoU &
  0.5392 &
  0.5527 &
  0.2012 &
  0.2101 &
  0.5985 &
  0.5985 &
  0.4463 &
  \textbf{0.4537} \\ \cline{2-10} 
 &
  Acc &
  0.7297 &
  0.7640 &
  0.3700 &
  0.4312 &
  0.7456 &
  0.7395 &
  0.7240 &
  \textbf{0.7303} \\ \hline
\multirow{2}{*}{\begin{tabular}[c]{@{}c@{}}512X512\\ FCN\end{tabular}} &
  IoU &
  0.4907 &
  0.4954 &
  0.1982 &
  0.2034 &
  0.5952 &
  0.5941 &
  0.4281 &
  \textbf{0.4309} \\ \cline{2-10} 
 &
  Acc &
  0.7159 &
  0.7138 &
  0.3753 &
  0.3930 &
  0.7491 &
  0.7525 &
  0.7236 &
  \textbf{0.7265} \\ \hline

\end{tabular}
\end{table*}

Here,  we describe  the details of our proposed method. We first divide the full image into non-overlapping square patches. The objective is to  segment each patch. However, instead of segmenting each patch independently, we propose to capture the context information held by the patches that are immediate neighbors. Let, the patch we are going to segment be denoted as I. The patch I has eight neighbor patches. For the target patches that lie at borders of the full image, we have  used zero-valued patches to extrapolate the missing neighbors. 

At first, the  patch I is fed to the encoder. We get an output $ I_{e} \in R^{1\times C \times H \times W} $, which is the encoded version of patch I. $I_e$ is then reshaped to $1 \times C \times HW$. 
After that, the neighbors are put into the encoder keeping the gradients of encoder weights and biases frozen, i.e., the weights and biases of encoder are not updated during training for this particular step. 
The output $ N_e \in R ^ {8 \times C \times H \times W} $ is later reshaped to $8 \times C \times HW $. We then concatenate $ I_e$ to $ N_e$ because patch $I$ needs to get a global view of the neighboring patches, as well as the whole patch of itself. Now, to calculate the feature interdependencies between $N_{e}$ and $I_{e}$, we multiply $ I_e^T \in R ^{1 \times HW \times C}$ with $ N_e$ and then apply softmax to the last axis to get the weight matrix $ W_c \in R ^ {9 \times HW \times HW} $. This weight matrix is significant for our objective. It consists of 9 matrices. Let, each element of $W_c$ be denoted as $w_{ji_{k}}$; where $k \in 1,2 ... 9$. Then, $w_{ji_{k}}$ measures the impact of $i^{th}$ position of $N_{e_{k}}$ on $ j^{th}$ position of $I_e$. More similar feature representations of the two positions contribute to greater correlation between them. These weight values would act like gates that control the flow of spatial contextual information coming from the neighbors.

Now, we need to fuse contextual information held by the neighboring patches to $I_{e}$. To do that, we first multiply $N_e$ with the transpose of weight matrix to get weighted neighbors. Then, we reshape the weighted neighbors to a size of $9 \times C \times H \times W $ to get the final weighted neighbors $ N_w$. Each pixel position for each neighbor  $ N_{w_{k}}$ is a linear combination of rest of the pixel positions from that particular neighbor $N_{w_{k}}$. The weights of the linear combination is the weight values from $ W_c$. 

In the decoding procedure, $ N_w$ is first summed through it's batch dimension. $ N_w$ is then elementwise multiplied with a learnable parameter $ \alpha$ of size $H \times W $, which is initialized with a value of 0. After that, element wise summation is performed between $ I_e$ and $ N_w$ to get $D_e$. This way, each pixel position of $ D_e$ has the information from patch I and the contextual information from all the neighbors surrounding the patch I. So, $ D_e$ now has a global view of the whole 9 patches. Finally, $ D_e$ is put into the decoder, where it is upsampled to the original resolution of the  patch I.   
\section{Experiments}
\label{sec:typestyle}
Our experimental objective is composed of two parts. At first, we compare our method against basic patch-based approach using baseline segmentation architectures to show the efficacy of our proposed framework. We use two datasets, 1) JSRT Chest X-ray  \cite{shiraishi2000development} and 2) Dhaka Satellite Data (DSD). Next, we observe how our method advances the current state of the art results on GID Satellite Dataset \cite{tong2020land}. We compare our method against their multi-patch based method. All experiments are performed using Pytorch implementation. 
\begin{table*}[ht!]
\footnotesize 
\centering
\caption{Baseline Comparison for Dhaka Satellite Dataset} 
\label{tab:Dhaka}
\begin{tabular}{|c|l|l|l|l|l|l|l|l|l|l|l|l|l|}
\hline
\multicolumn{2}{|l|}{\multirow{2}{*}{}} &
  \multicolumn{2}{c|}{Forest} &
  \multicolumn{2}{c|}{Built-up} &
  \multicolumn{2}{c|}{Water} &
  \multicolumn{2}{c|}{Farmland} &
  \multicolumn{2}{c|}{Meadow} &
  \multicolumn{2}{c|}{mIoU/OA} \\ \cline{3-14} 
\multicolumn{2}{|l|}{} &
  \multicolumn{1}{c|}{w/o} &
  \multicolumn{1}{c|}{w} &
  \multicolumn{1}{c|}{w/o} &
  \multicolumn{1}{c|}{w} &
  \multicolumn{1}{c|}{w/o} &
  \multicolumn{1}{c|}{w} &
  \multicolumn{1}{c|}{w/o} &
  \multicolumn{1}{c|}{w} &
  \multicolumn{1}{c|}{w/o} &
  \multicolumn{1}{c|}{w} &
  \multicolumn{1}{c|}{w/o} &
  \multicolumn{1}{c|}{w} \\ \hline
\multirow{2}{*}{\begin{tabular}[c]{@{}c@{}}56X56\\ Deeplab\end{tabular}} &
  IoU &
  0.3783 &
  0.3816 &
  0.5673 &
  0.5861 &
  0.4572 &
  0.4646 &
  0.5700 &
  0.5849 &
  0.2451 &
  0.2700 &
  0.4436 &
  \textbf{0.4574} \\ \cline{2-14} 
 &
  Acc &
  0.5790 &
  0.5561 &
  0.6945 &
  0.7243 &
  0.6801 &
  0.7254 &
  0.7265 &
  0.7292 &
  0.3992 &
  0.4306 &
  0.6499 &
  \textbf{0.6651} \\ \hline
\multirow{2}{*}{\begin{tabular}[c]{@{}c@{}}56X56\\ FCN\end{tabular}} &
  IoU &
  0.3168 &
  0.3319 &
  0.5315 &
  0.5346 &
  0.3729 &
  0.3830 &
  0.5178 &
  0.5349 &
  0.1845 &
  0.1627 &
  0.3847 &
  \textbf{0.3894} \\ \cline{2-14} 
 &
  Acc &
  0.6722 &
  0.6187 &
  0.6419 &
  0.6457 &
  0.5645 &
  0.6432 &
  0.6767 &
  0.7248 &
  0.2626 &
  0.2240 &
  0.6005 &
  \textbf{0.6114} \\ \hline
\multirow{2}{*}{\begin{tabular}[c]{@{}c@{}}112X112\\ Deeplab\end{tabular}} &
  IoU &
  0.3299 &
  0.3774 &
  0.5498 &
  0.5645 &
  0.4274 &
  0.4426 &
  0.5659 &
  0.5395 &
  0.2502 &
  0.3187 &
  0.4246 &
  \textbf{0.4486} \\ \cline{2-14} 
 &
  Acc &
  0.5246 &
  0.6418 &
  0.7132 &
  0.6707 &
  0.6292 &
  0.6013 &
  0.6948 &
  0.6052 &
  0.4061 &
  0.6648 &
  0.6365 &
  \textbf{0.6394} \\ \hline
\multirow{2}{*}{\begin{tabular}[c]{@{}c@{}}112X112\\ FCN\end{tabular}} &
  IoU &
  0.3260 &
  0.3282 &
  0.4888 &
  0.5495 &
  0.3705 &
  0.3819 &
  0.4843 &
  0.5413 &
  0.2513 &
  0.2110 &
  0.3842 &
  \textbf{0.4024} \\ \cline{2-14} 
 &
  Acc &
  0.7173 &
  0.6791 &
  0.5524 &
  0.6643 &
  0.7553 &
  0.5559 &
  0.5675 &
  0.6889 &
  0.4608 &
  0.3063 &
  0.5764 &
  \textbf{0.6186} \\ \hline
\multirow{2}{*}{\begin{tabular}[c]{@{}c@{}}224X224\\ Deeplab\end{tabular}} &
  IoU &
  0.3590 &
  0.3672 &
  0.5869 &
  0.5859 &
  0.3817 &
  0.4404 &
  0.5569 &
  0.5643 &
  0.2556 &
  0.2899 &
  0.4280 &
  \textbf{0.4496} \\ \cline{2-14} 
 &
  Acc &
  0.6331 &
  0.6593 &
  0.7296 &
  0.7385 &
  0.6404 &
  0.6490 &
  0.6775 &
  0.6659 &
  0.3788 &
  0.4389 &
  0.6458 &
  \textbf{0.6566} \\ \hline
\multirow{2}{*}{\begin{tabular}[c]{@{}c@{}}224X224\\ FCN\end{tabular}} &
  IoU &
  0.3489 &
  0.3581 &
  0.5264 &
  0.5559 &
  0.3969 &
  0.4025 &
  0.5550 &
  0.5559 &
  0.2508 &
  0.2591 &
  0.4156 &
  \textbf{0.4263} \\ \cline{2-14} 
 &
  Acc &
  0.7389 &
  0.6778 &
  0.5967 &
  0.6544 &
  0.6215 &
  0.7151 &
  0.6939 &
  0.6868 &
  0.3962 &
  0.3941 &
  0.6214 &
  \textbf{0.6359} \\ \hline
\multirow{2}{*}{\begin{tabular}[c]{@{}c@{}}448X448\\ Deeplab\end{tabular}} &
  IoU &
  0.3449 &
  0.3577 &
  0.5452 &
  0.5598 &
  0.4065 &
  0.4443 &
  0.4826 &
  0.5591 &
  0.2477 &
  0.2720 &
  0.4054 &
  \textbf{0.4386} \\ \cline{2-14} 
 &
  Acc &
  0.7238 &
  0.6307 &
  0.6590 &
  0.6934 &
  0.6565 &
  0.6836 &
  0.5701 &
  0.6503 &
  0.4222 &
  0.4711 &
  0.6025 &
  \textbf{0.6393} \\ \hline
\multirow{2}{*}{\begin{tabular}[c]{@{}c@{}}448X448\\ FCN\end{tabular}} &
  IoU &
  0.3365 &
  0.3431 &
  0.5560 &
  0.5607 &
  0.3999 &
  0.4005 &
  0.5340 &
  0.5214 &
  0.2415 &
  0.2791 &
  0.4136 &
  \textbf{0.4209} \\ \cline{2-14} 
 &
  Acc &
  0.5474 &
  0.6041 &
  0.6859 &
  0.7031 &
  0.7447 &
  0.7247 &
  0.6542 &
  0.5947 &
  0.4074 &
  0.4896 &
  0.6228 &
  \textbf{0.6247} \\ \hline

\end{tabular}
\end{table*}

\textbf{Performance Metrics:} We report our performance using intersection over union (IoU), mean intersection over union (mIoU), class specific accuracy (Acc) and overall accuracy (OA) \cite{tong2020land}. IoU is defined as 

$$ IoU_j = \frac{\sum_{i=1}^{n}TP_{ij}}{\sum_{i=1}^{n}(TP_{ij}+FP_{ij}+FN_{ij})}   $$

$TP_{ij}$, $FP_{ij}$, and $FN_{ij}$ are correspondingly the number of pixels in image $i$, which are correctly predicted as class $j$, incorrectly predicted as class $j$, incorrectly predicted as any other class than class $j$, respectively. The mIoU is defined as the average IoU among all classes.

Let $ P_{ab}$ denote the number of pixels of class a predicted to belong to class b, and let $t_a = \sum_{b}^{}P_{ab}$ be the total number of pixels belonging to class $a$. Let, $t_b = \sum_{a}^{}P_{ab}$ be the total number of pixels predicted to class $b$. Class specific accuracy is defined as the percentage of correctly classified pixels for each class. Overall Accuracy (OA) is the percentage of correctly classified pixels and all pixels in the entire image.

\[ Accuracy = \frac{P_{aa}}{t_a} \; \; \; \; ; \; \; \; \; OA = \frac{\sum_{a}^{}P_{aa}}{\sum_{a}^{}t_a}    \]


In all the experiments the background class is discarded during the calculation of the metrics.

\textbf{Comparing Baselines:}
Here, we compare our method against basic patch based approach. We first use the JSRT Chest X-ray  dataset \cite{shiraishi2000development}. Segmentation in Chest Radiographs \cite{van2006segmentation} contains manual segmentation for lung fields, heart, and clavicles of JSRT Dataset. The images are 1024×1024 in resolution. We show the comparative results with different patch sizes to demonstrate the significance of our framework.  Since this dataset does not have any explicit test set, we have performed each experiment with three fold cross validation. Here, we report the average scores of the three runs. We use two different segmentation architectures: FCN-32 \cite{long2015fully} and Deeplab v3+ \cite{chen2018encoder}. We select a comparatively older and a comparatively newer architecture to verify the effectiveness of our method. Deeplab v3+ has explicit encoder and decoder blocks. For FCN-32, we use the only upsampling layer as decoder and the rest as encoder. We perform the experiment with Adam optimizer using a learning rate of 0.0001. For the first three patch sizes (64×64, 128×128, 256×256), we use a batch size of 32. For 512×512 patch size we use batchsize of 8 due to memory constraint. We run the training for 30 epochs. %
The results are reported in table \ref{tab:JSRT}. It is easy to note that both the FCN 32 and Deep Lab v3+ work well with our proposed context aware framework. The mIoU and overall accuracy values indicate the superior performance of our proposed method.

Next, we show baseline comparison for Dhaka Satellite Dataset (DSD). It has a single image with resolution 51146×15233. To increase diversity of our experiment, we opt for four different patch sizes than JSRT experiment. DSD has an explicit test set. We use both FCN-32 and Deeplab v3+ for this experiment with Adam optimizer with a learning rate of 0.0001. We run each experiment for 20 epochs. For both JSRT and DSD, we use pixel-wise cross-entropy loss. From table \ref{tab:Dhaka}, it is easy to notice that the IoUs are significantly better with our proposed approach than the baseline models for five classes for most of the experiments. The mIoU and OAs are always improved with the proposed framework. 

For both JSRT and DSD datasets, mIoU and OA increase for all patch sizes using our proposed framework. However, for DSD, the improvements are more prominent because DSD has much higher resolution images than JSRT. Therefore, it gets more advantage from the contextual information. One significant observation is that patch based methods struggle when the patch size is very small compared to the full image. The reason is that the ratio between total number of border pixels and non-border pixels is high for smaller patches. So, for patch based method, the errors at border pixels add up, resulting in performance drop. However, proposed context aware framework does not suffer from this limitation; hence it achieves better segmentation for smaller patches. 

\begin{table}[!h]
\centering
\caption{Comparison with GID (Classwise Accuracy)} 
\label{tab:GID}
\footnotesize
\begin{tabular}{|l|c|c|c|c|c|}
\hline
Methods    & \multicolumn{1}{l|}{Built-up} & \multicolumn{1}{l|}{Farmland} & \multicolumn{1}{l|}{Forest} & \multicolumn{1}{l|}{Meadow} & \multicolumn{1}{l|}{Water} \\ \hline
MLC+Fusion   & 61.81 & 67.38 & 36.15 & 2.92           & 82.09 \\ \hline
RF+Fusion    & 62.61 & 71.12 & 36.10 & 3.94           & 84.29 \\ \hline
SVM+Fusion   & 61.28 & 72.27 & 23.01 & 2.26           & 54.18 \\ \hline
MLP+Fusion & 58.69 & 72.40 & 32.86 & 2.67           & 83.99 \\ \hline
PT-GID       & 88.42 & 91.85 & 79.42 & \textbf{70.55} & 87.60 \\ \hline
Our Method & \textbf{97.57}                & \textbf{92.97}                & \textbf{84.64}              & 59.12                       & \textbf{95.03}             \\ \hline
\end{tabular}
\end{table}


\textbf{Comparing with State of the Art:}
We now validate our proposed framework on GID satellite dataset \cite{tong2020land}. GID has a total of 150 very high resolution satellite images, each of them has a size of 7200×6800. Because of such wide variety of high resolution images, this dataset is very much suitable for our experiment. The authors divided the dataset into 120 train images and 30 test images. We further divide the train set into 80\% train and 20\% validation set. We use Deeplab v3+ as our segmentation architecture and a patch size of 224×224 as it showed the best performance on DSD. We train for 15 epochs with a learning rate of 0.0001 using pixel-wise cross-entropy loss. In table \ref{tab:GID}, we show our result. The results of the competing methods are directly taken from \cite{tong2020land}. The performance metric is the percentage class wise accuracy as used in \cite{tong2020land}. It is observed that our method advances the current state of the art accuracy on four out of five classes. 
Accuracy on meadow class is less because it has the least number of pixels in the training set and we did not perform any data augmentation.

\vspace{-5 pt}
\section{Conclusion}
\vspace{-5 pt}
In this paper, we have presented a novel framework that fuses contextual information from neighboring patches to aid in segmentation. This way, the network has a wider view of the target patch and becomes more confident on the segmentation task, which is reflected in our diverse experiments. Our framework has also advanced the state of the art accuracy on GID dataset. 

\section{Acknowledgment}
This project is supported by ICT Division - Government of Bangladesh and Independent University Bangladesh (IUB).



\bibliographystyle{IEEEbib}
\bibliography{ICIP}

\end{document}